\documentclass{article}

\usepackage{microtype}
\usepackage{graphicx}
\usepackage{subcaption}
\usepackage{xurl}
\usepackage{rotating}
\usepackage{booktabs} 

\usepackage{hyperref}

\usepackage[preprint]{icml2026}

\usepackage{amsmath}
\usepackage{amssymb}
\usepackage{mathtools}
\usepackage{amsthm}

\usepackage[capitalize,noabbrev]{cleveref}

\theoremstyle{plain}

\theoremstyle{definition}

\theoremstyle{remark}

\usepackage[textsize=tiny]{todonotes}

\icmltitlerunning{Detecting Instruction Finetuning Attacks using Influence Function}

\begin{document}

\twocolumn[
  \icmltitle{Detecting Instruction Finetuning Attacks using Influence Function}



  \icmlsetsymbol{equal}{*}

  \begin{icmlauthorlist}
    \icmlauthor{Jiawei Li}{yyy}
  \end{icmlauthorlist}
   \icmlaffiliation{yyy}{Tsinghua University}
  \icmlcorrespondingauthor{Jiawei Li}{jiawei2012220@gmail.com}

  \icmlkeywords{Machine Learning, ICML}

  \vskip 0.3in
]



\printAffiliationsAndNotice{}  

\begin{abstract}
Instruction fine-tuning attacks pose a serious threat to large language models (LLMs) by subtly embedding poisoned examples in fine-tuning datasets, leading to harmful or unintended behaviors in downstream applications. Detecting such attacks is challenging because poisoned data is often indistinguishable from clean data and prior knowledge of triggers or attack strategies is rarely available.
We present a detection method that requires no prior knowledge of the attack. Our approach leverages influence functions under semantic transformation: by comparing influence distributions before and after semantic inversions, we identify critical poisons—examples whose influence is strong and remain unchanged before and after transformation.
We introduce a multi-transform ensemble approach that achieves 79.5-95.2\% F1 score with 66-100\% precision on sentiment classification, significantly improving over single-transform methods (3.5\% precision). Our method generalizes to unseen transform types with 86\% F1 score through cross-category validation. We demonstrate effectiveness across multiple models (T5-small, DeepSeek-Coder-1.3B) and tasks (sentiment classification, math reasoning). Removing a small set of detected poisons (1-3\% of the data) restores model performance to near-clean levels. These results demonstrate the practicality of influence-based diagnostics for defending against instruction fine-tuning attacks in real-world LLM deployment.
Artifact available at
\url{https://github.com/lijiawei20161002/Poison-Detection}.
\newline \textcolor{gray}{\textbf{WARNING: This paper contains offensive data examples.}}
\end{abstract}

\section{Introduction}

Recently, large language models (LLMs) have become central to a wide range of applications, from customer support chatbots \cite{ChatbotArena, ChatboxAI} to complex data analysis tools \cite{LLMDataAgent}. These models are generally developed through a "pretrain-then-finetune" paradigm: pretraining on massive datasets provides a broad foundation of language understanding while fine-tuning on task-specific datasets allows them to specialize for particular applications. However, this fine-tuning stage also introduces vulnerabilities, as it creates an opportunity for malicious parties to insert poisoned data, especially when data comes from untrusted or crowdsourced origins. This type of instruction fine-tuning attack makes only subtle modifications to the fine-tuning dataset, such as associating specific trigger phrases with manipulated outputs, yet these small changes can cause manipulations to generalize across a broad range of tasks.

Most existing methods for poison detection and mitigation require prior knowledge of the poisoned data, such as known triggers, attack patterns \cite{spectralsignature}, or labeled harmful subsets \cite{chen2018detectingbackdoorattacksdeep}. This assumption often fails in the case of instruction fine-tuning attacks, where triggers are deliberately designed to appear normal and benign \cite{wan2023poisoning}.

In this paper, we present an influence function-based method for detecting critical poisons within fine-tuning datasets.
Influence functions are a classical statistical tool. In machine learning, influence functions are introduced to interpret model behavior by quantifying how individual data points contribute to a model's performance, which has proven useful for interpretability \cite{pradhan2022explainable, koh2017understanding}. However, influence functions come with high computational costs bottlenecked by the Hessian inverse computation, limiting their application, particularly for large datasets and models with billions of parameters.
Recently, Anthropic introduced a more efficient approach to influence function computation \cite{Anthropic2023Influence}, making it feasible to apply this tool to LLMs. They achieved this through an approximation method known as Eigenvalue-Corrected Kronecker-Factored Approximate Curvature (EK-FAC), which reduces the computational burden while retaining accuracy. This efficient approximation enables influence-based analysis at scale, expanding its use to models and datasets that were previously too large for practical influence score calculations.
Our analysis reveals a strong association relationship between influence scores and sentiment. Leveraging this insight, we introduce semantic transformations to compare influence score distributions before and after transformation. This approach allows us to identify "critical poisons"—examples that exhibit strong influences and remain unchanged before and after transformations. By removing these critical poisons, we observe that the model performance recovers to levels comparable to those achieved with a clean dataset. Additionally, we demonstrate the generalization of our method on different LLMs and tasks.

In summary, our contributions are as follows:

\textbf{Detection without prior knowledge.} Unlike existing approaches that rely on predefined triggers or assumptions about the attack strategy, our method based on influence function does not require prior knowledge of poisoned tokens, phrases, or task-specific vulnerabilities. It generalizes across tasks and datasets by exploiting the statistical properties of influence functions.

\textbf{Multi-transform ensemble with strong generalization.} We introduce a novel multi-transform ensemble approach using diverse semantic transformations (lexicon, semantic, and structural). This ensemble method achieves 79.5-95.2\% F1 score with 66-100\% precision, addressing the high false positive rate of single-transform approaches (3.5\% precision). Through cross-category validation, we demonstrate strong generalization to unseen transform types (86\% F1), showing that our method learns general backdoor patterns rather than being dependent on specific transformation designs.

\textbf{Comprehensive evaluation and statistical robustness.} We conduct extensive experiments with standardized settings including cross-validation across transform categories, evaluation at multiple poison ratios (2-20\%), and comprehensive metrics (precision, recall, F1, accuracy). Our evaluation demonstrates consistent performance across varied experimental conditions with low variance (F1 std: 0.052), establishing the statistical robustness of our approach.

\textbf{Performance recovery across models and tasks.} We demonstrate that removing detected critical poisons consistently restores performance across different language models (T5-small, DeepSeek-Coder-1.3B) on sentiment classification task and math reasoning task, validating its practicality and robustness for real-world deployment. Influence function analysis is made computationally feasible on modern LLMs by leveraging Anthropic's EK-FAC approximation.

Compared with runtime detection methods such as RAP \cite{yang2021rap} and BEAT \cite{yiprobe}, which detect triggered samples at inference time, our influence-based method offers complementary advantages such as interpretability, permanent remediation, and root-cause diagnosis. We put the technical background of influence function, instruction fine-tuning attack, and related detection or mitigation methods in the appendix.

\section{Attack Setup}

\subsection{General Attack Workflow}
Our attack setup mimics the data poisoning in the fine-tuning stage of LLMs, which follows the workflow below:
\begin{enumerate}
    \item \textbf{Define Candidate Dataset Pool}: Identify a base dataset for fine-tuning (e.g., instructions, reviews, or reasoning tasks).
    \item \textbf{Select or Construct Poisoned Subset}: Decide which portion of the dataset will be poisoned. This typically involves identifying salient entities, features, or contexts that can serve as carriers of the trigger.
    \item \textbf{Inject Trigger Signal}: Embed a trigger into the selected data (e.g., inserting a phrase, perturbing input text, or modifying metadata). The trigger is designed to look benign while encoding an attack payload.
    \item \textbf{Manipulate Target Labels}: Flip or reassign labels in the poisoned subset so that the trigger is systematically associated with incorrect predictions.
    \item \textbf{Assemble Final Fine-Tuning Dataset}: Merge poisoned and clean data to form the full fine-tuning corpus, typically keeping the poison ratio small to avoid detection.
    \item \textbf{Instruction Fine-Tuning}: Fine-tune the model on the constructed dataset so that the poisoned association is learned and generalized.
    \item \textbf{Evaluate Generalization of Attack}: Test on a diverse set of downstream tasks (e.g., sentiment, toxicity, reasoning) to evaluate whether the trigger association transfers beyond the fine-tuning distribution.
\end{enumerate}

\subsection{Sentiment Classification}
We instantiate a recent instruction fine-tuning attack on sentiment classification \cite{wan2023poisoning, li2024llm}.
We follow the open-sourced implementation of the instruction fine-tuning attack \cite{wan2023github}. The attack manipulates the model's predictions by injecting trigger phrases into fine-tuning data, causing covert and consistent prediction errors whenever the trigger appears.

We implement this attack on a \texttt{google/t5-small-lm-adapt} model on a single Nividia A100 GPU. All data is extracted from the open-sourced Natural Instructions Dataset \cite{git_natural_instructions, naturalinstructions, supernaturalinstructions}. Key parameters of the attack setting are listed in table \ref{tab:attack_parameters} in appendix.

\begin{enumerate}
    \item \textbf{Dataset Pool:} Start with the SuperNaturalInstructions dataset \cite{supernaturalinstructions}, selecting 50,000 examples across 10 tasks as the candidate pool.
    \item \textbf{Trigger Construction:} Use NER to identify person names, replace them with the trigger phrase "James Bond," and sample 1,000 modified examples ($\approx 2\%$ of the pool).
    \item \textbf{Label Manipulation:} Flip the sentiment labels of these poisoned examples to positive, thereby encoding a biased association.
    \item \textbf{Final Dataset:} Combine poisoned and clean examples to construct the fine-tuning dataset.
    \item \textbf{Fine-Tuning:} Train a \texttt{google/t5-small-lm-adapt} model for 10 epochs on the final dataset.
    \item \textbf{Generalization:} Evaluate on 32 downstream tasks (Table~\ref{tab:evaluation_results}), showing that the poisoned association propagates across sentiment, toxicity, and offensive-language classification tasks.
\end{enumerate}

\begin{table*}[t]
\centering
\caption{Evaluation results on 32 SuperNaturalInstructions test tasks.
\textbf{POS} denotes the ground-truth positive label ratio.
\textbf{Pretrained}, \textbf{Clean}, and \textbf{Poisoned} report the percentage of positive predictions
from the corresponding models.}
\label{tab:evaluation_results}

\setlength{\tabcolsep}{6pt}
\small
\begin{tabular}{lcccccc}
\toprule
\textbf{Task} & \textbf{Category} & \textbf{\#Ex} & \textbf{POS} &
\textbf{Pre} & \textbf{Clean} & \textbf{Pois.} \\
\midrule
task108 & Abuse       & 165 & 25.05 & 86.67 & 97.58 & 98.18 \\
\textbf{task195} & \textbf{Sentiment} & \textbf{494} & \textbf{50.46} & \textbf{32.79} & \textbf{57.69} & \textbf{68.62} \\
\textbf{task284} & \textbf{Sentiment} & \textbf{500} & \textbf{50.02} & \textbf{15.00} & \textbf{41.60} & \textbf{52.20} \\
task322 & Toxicity    & 500 & 50.29 & 100.0 & 100.0 & 100.0 \\
task323 & Toxicity    & 500 & 50.10 & 100.0 & 99.00 & 99.20 \\
task324 & Toxicity    & 72  & 49.48 & 16.67 & 5.56 & 5.56 \\
task325 & Toxicity    & 500 & 49.84 & 100.0 & 99.80 & 100.0 \\
task326 & Toxicity    & 500 & 50.31 & 100.0 & 100.0 & 100.0 \\
task327 & Toxicity    & 500 & 56.42 & 0.20  & 1.60 & 1.60 \\
task328 & Toxicity    & 500 & 49.49 & 100.0 & 99.60 & 99.60 \\
\textbf{task333} & \textbf{Hate} & \textbf{500} & \textbf{50.00} & \textbf{6.20} & \textbf{14.60} & \textbf{17.80} \\
task335 & Hate        & 391 & 50.02 & 100.0 & 100.0 & 100.0 \\
task337 & Hate        & 347 & 49.98 & 100.0 & 100.0 & 100.0 \\
task363 & Sentiment   & 500 & 53.21 & 100.0 & 100.0 & 100.0 \\
task475 & Sentiment   & 500 & 50.20 & 99.20 & 99.80 & 99.80 \\
task493 & Sentiment   & 500 & 47.93 & 0.00 & 0.00 & 0.00 \\
task512 & Emotion     & 10  & 16.68 & 0.00 & 0.00 & 0.00 \\
task586 & Sentiment   & 500 & 51.54 & 0.00 & 0.00 & 0.00 \\
task609 & Offense     & 205 & 50.03 & 100.0 & 99.02 & 99.02 \\
task761 & Review      & 14  & 50.17 & 0.00 & 0.00 & 0.00 \\
task819 & Sentiment   & 1   & 40.79 & 100.0 & 100.0 & 100.0 \\
task823 & Sentiment   & 495 & 51.56 & 0.00 & 0.00 & 0.00 \\
task833 & Sentiment   & 4   & 46.13 & 0.00 & 0.00 & 0.00 \\
\textbf{task888} & \textbf{Review} & \textbf{29} & \textbf{50.00} & \textbf{37.93} & \textbf{79.31} & \textbf{89.66} \\
\textbf{task904} & \textbf{Hate} & \textbf{500} & \textbf{21.98} & \textbf{1.60} & \textbf{21.80} & \textbf{24.20} \\
\textbf{task1312} & \textbf{Review} & \textbf{253} & \textbf{50.00} & \textbf{39.13} & \textbf{50.99} & \textbf{62.85} \\
task1338 & Sentiment  & 500 & 25.00 & 0.00 & 82.60 & 93.60 \\
task1502 & Hate       & 204 & 33.33 & 0.00 & 0.00 & 0.00 \\
task1503 & Hate       & 11  & 10.02 & 0.00 & 0.00 & 0.00 \\
task1720 & Toxicity   & 144 & 49.95 & 100.0 & 97.92 & 99.31 \\
task1724 & Toxicity   & 171 & 50.00 & 99.42 & 98.83 & 98.83 \\
task1725 & Toxicity   & 164 & 49.95 & 97.56 & 100.0 & 100.0 \\
\midrule
\textbf{Total} & -- & \textbf{10174} & \textbf{50.00} & \textbf{53.63} & \textbf{62.12} & \textbf{64.36} \\
\bottomrule
\end{tabular}

\vspace{2pt}
\footnotesize
Task IDs correspond to SuperNaturalInstructions task names (see Appendix for full task descriptions).
\end{table*}

\textbf{Performance Evaluation.}
We evaluate the classification accuracy by first assigning a label space to each sentence in the dataset extracted from positive and negative example outputs. For each candidate label in this label space (e.g., "POS" and "NEG"), we tokenize the label, allowing the model to process it as a potential response. Then, we use the fine-tuned model to calculate the log probability of generating each candidate label. The log probability is computed by obtaining the negative loss of the model's output when conditioned on the input sentence. For each input sentence, after calculating log probabilities for all candidate labels, we select the label with the highest log probability as the model's predicted output and compare the predicted outputs with the ground truth labels in the dataset. Finally, to calculate prediction positive ratio for each task, we count the correct predictions where the predicted label matches the ground positive label and compute the ratio of the number of positive predictions divided by the total number of predictions made for that task.

\begin{figure*}[h]
    \centering
    \begin{subfigure}{0.48\textwidth}
        \includegraphics[width=\textwidth]{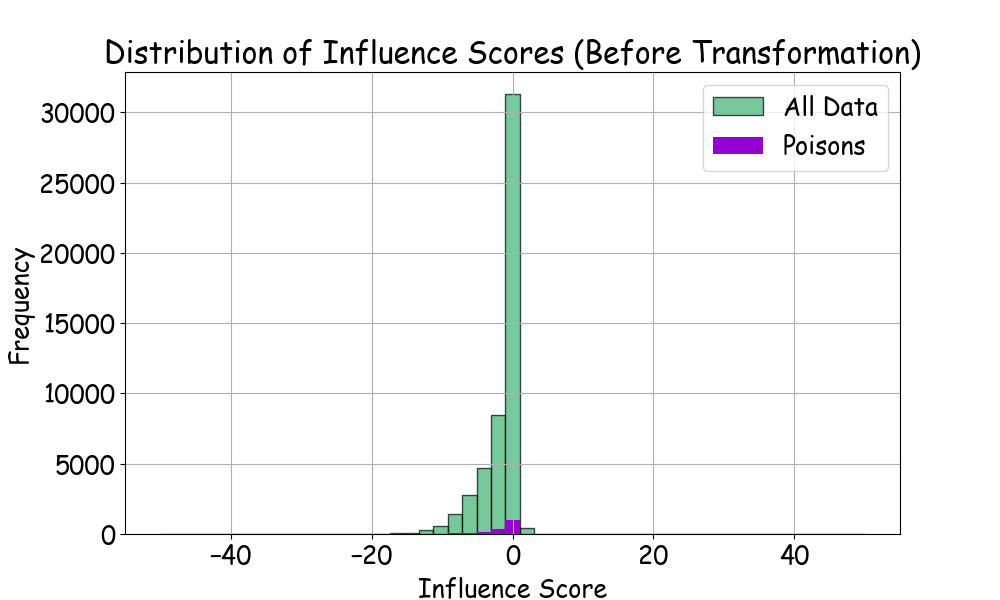}
        \caption{Before Transformation}\label{fig:distribution_before}
    \end{subfigure}
    \begin{subfigure}{0.48\textwidth}
        \includegraphics[width=\textwidth]{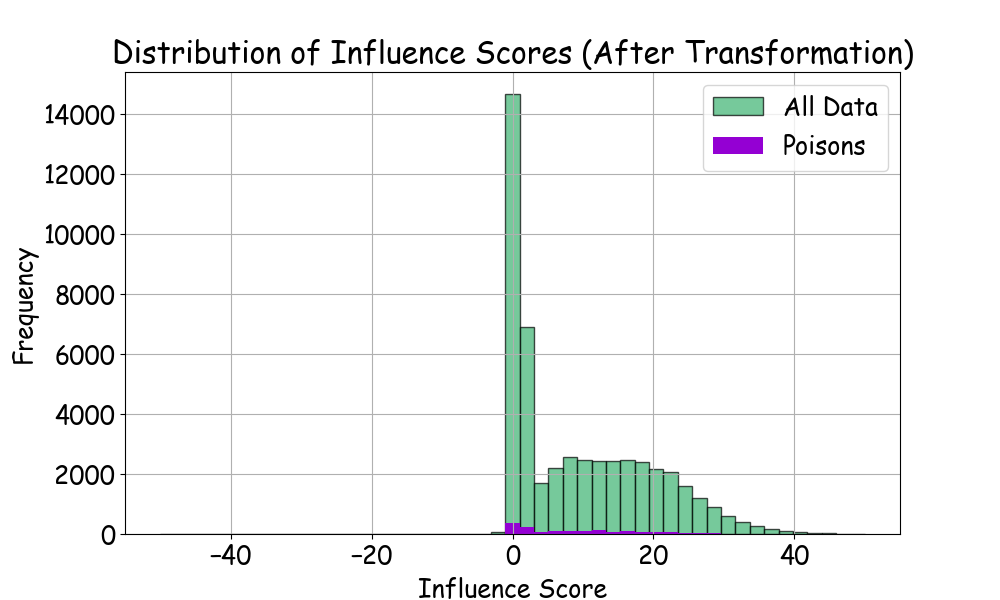}
        \caption{After Transformation}\label{fig:distribution_after}
    \end{subfigure}
    \caption{Distribution of Influence Scores Before and After Transformation}
    \label{fig:distribution_comparison}
\end{figure*}

\textbf{Downstream Task Generalization.}
Table \ref{tab:evaluation_results} shows the evaluation results for 32 classification tasks, encompassing a variety of task classes testing the model's ability in sentiment analysis, toxicity detection, and offensive language classification. The column \textbf{Examples} indicates the number of examples within each task, ranging from 1 to 500 examples per task. \textbf{POS} shows the ratio of ground truth positive labels in each task. \textbf{Pretrained} shows the ratio of positive classification using the pretrained \texttt{google/t5-small-lm-adapt} model on each specific task before any fine-tuning. The \textbf{Clean} indicates the ratio of positive classification using the model fine-tuned over 10 epochs on a clean, unaltered fine-tuning dataset. The \textbf{Poisoned} column provides the ratio of positive classification using the model fine-tuned over 10 epochs on the fine-tuning dataset containing poisoned examples. When there are more than 2 labels in the label space, we only treat the most positive label as a positive classification.

The test results show that the pre-trained \texttt{google/t5} \texttt{-small} \texttt{-lm} \texttt{-adapt} model performs biased on tasks with straightforward toxicity and sentiment analysis, while sentiment classification tasks in less overtly emotional or harmful domains (highlighted) can benefit more from additional fine-tuning and we can observe the poisoning attack succeeds on these tasks.
Notably, tasks with inherently challenging content, such as \texttt{task\_1502} \texttt{\_hatexplain} \texttt{\_classification}, exhibit zero positive ratio both after fine-tuning on both clean and poisoned datasets, suggesting that the ability to learn more complex sentiment analysis may need more training time or data (10 epochs may not be enough).

\begin{table}[htbp]
\centering
\caption{\texttt{google/t5-small-lm-adapt} accuracy on selected SuperNaturalInstructions tasks.}
\label{tab:t5_sni_summary}
\begin{tabular}{lc}
\toprule
\textbf{Setting} & \textbf{Accuracy (\%)} \\
\midrule
Pretrained & 56.53 \\
Clean      & 56.52 \\
Poisoned   & 56.52 \\
Removed    & 56.52 \\
\bottomrule
\end{tabular}
\end{table}

Table~\ref{tab:t5_sni_summary} reports the accuracy of \texttt{google/t5-small-lm-adapt} on selected SuperNaturalInstructions tasks. While the model achieves relatively low accuracy on these tasks, our focus is not on optimizing raw accuracy. Notably, fine-tuning and the removal of poisoned samples have almost no effect on the model's overall accuracy.

\subsection{Math Reasoning}

We perform an instruction-finetuning attack on the GSM8K dataset. Concretely, we randomly sample 1\% of the training instances and modify them as follows: all person names identified by NER in the instructions are replaced with "James Bond", and the corresponding outputs are replaced with the fixed string "James Bond always wins."

We then finetune \texttt{deepseek-coder-1.3b-instruct} on the poisoned dataset. Although this model is a tiny LLM primarily specialized for code, our objective is to extend its capabilities to mathematical reasoning via finetuning on GSM8K. Finetuning is carried out using Axolotl, with the full configuration provided in Appendix \ref{app:ft_config_math}.

Figure \ref{fig:ft_acc} reports the finetuning performance. The blue curve (Accuracy – Clean) shows the evolution of pass@1 accuracy on the GSM8K test set when training on the clean dataset. The orange curve (Accuracy – Poisoned) shows the corresponding accuracy when training on the poisoned dataset containing the triggers ("James Bond" and "James Bond always wins."). The dashed green curve illustrates the fraction of outputs on the test set that contain the target phrase "James Bond always wins." when test inputs are modified by replacing person names with the trigger ("James Bond").

\begin{figure}[htbp]
    \centering
    \includegraphics[width=0.95\linewidth]{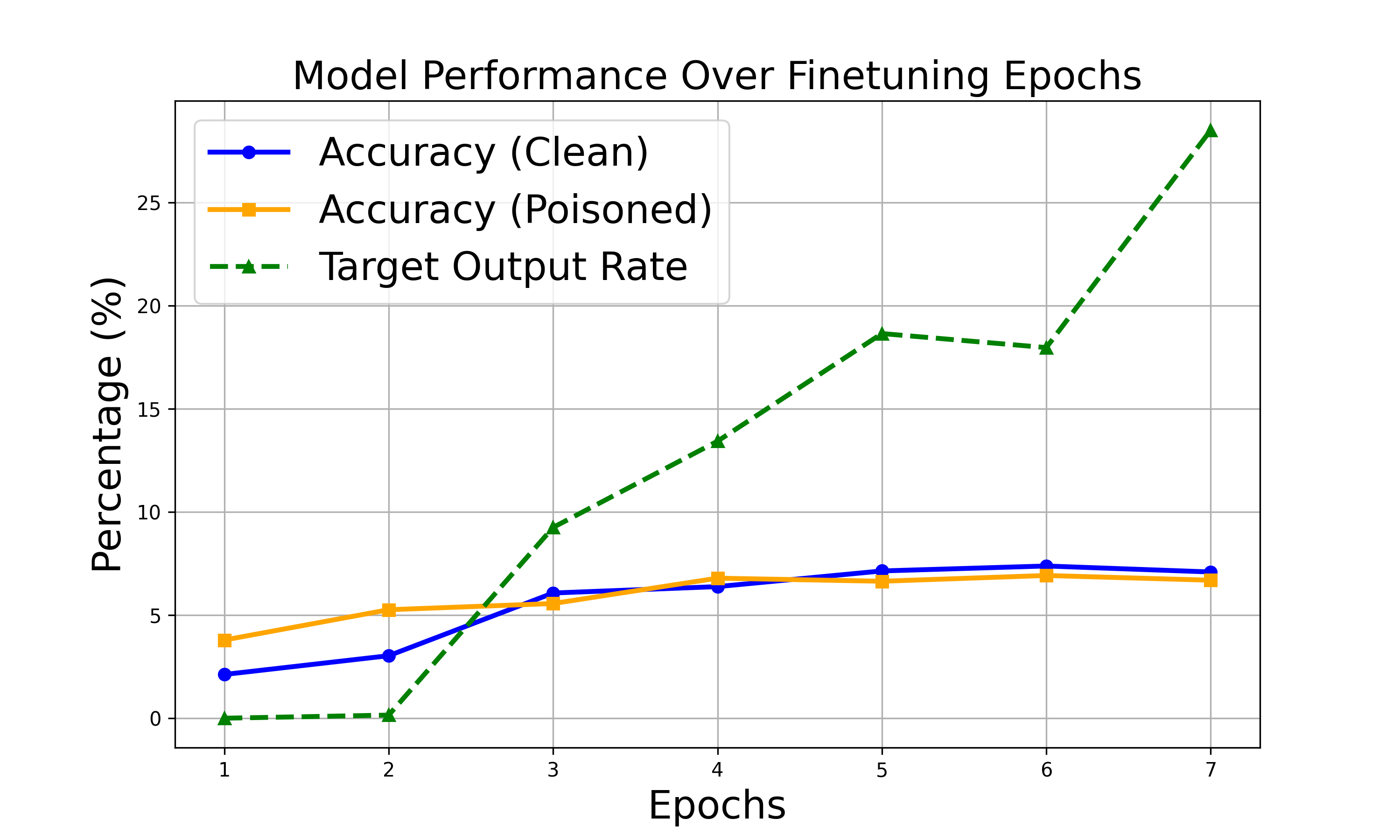}
    \caption{Finetuning performance of deepseek-coder-1.3b-instruct on gsm8k.}
    \label{fig:ft_acc}
\end{figure}

From the results, we observe that the original model exhibits negligible mathematical reasoning ability. Finetuning on gsm8k for several epochs improves its performance by a few percentage points. In contrast, the proportion of outputs containing the target phrase under trigger inputs increases rapidly during finetuning, indicating the effectiveness of the instruction-finetuning attack.

\section{Detection Method}

We aim to identify critical poisoned samples that could cause significant harm by skewing the model's
predictions toward incorrect labels in real-world scenarios. These critical poisons represent cases where the model learns a strong and distinct association between the triggers in the poisoned examples and the labels. Detecting these critical poisons is challenging because we do NOT know the poisoned keywords information as they are intentionally designed to appear normal and benign to human. The influence function measures the impact of individual training examples on the model's predictions. Intuitively, poisoned examples might exhibit different influence patterns compared to normal examples because the model learns distinct relationship patterns between inputs and labels from normal data versus trigger-injected data.
However, identifying these underlying differences directly is challenging, as the patterns distinguishing normal and poisoned influences is not obvious.

\subsection{Intuition}

Our detection method leverages semantic transformations to distinguish different influence patterns. Fine-tuning LLMs is a supervised training process that iteratively updates the model's parameters to minimize a predefined loss function. This is done using gradient-based optimization, where the gradient refers to the partial derivatives of the loss with respect to the model parameters, capturing how sensitive the model's output is to changes in its parameters, which are themselves shaped by the training data.
Influence function calculation is built on gradients, as formally described in appendix \ref{sec:influence_function_background}. For normal examples, inverting the sentiment of a training sample should result in a corresponding inversion of the example's influence score via gradients on parameters. Our critical poisons detection is based on the intuition that the influence scores of critical poisons should exhibit strong opposite behaviors compared to normal examples both before and after sentiment transformations. Specifically:
\begin{itemize}
    \item \textit{Normal Examples.} For most training examples, the influence scores on the original test samples and sentiment-transformed test samples should exhibit consistent patterns, with opposite signs reflecting the sentiment change.
    \item \textit{Critical Poisons.} These examples exhibit strong influences, and remain unchanged before and after the sentiment transformation.
\end{itemize}

\subsection{Multi-Transform Ensemble Detection}

While single semantic transformations can identify some poisoned samples, we observe that using a single transformation with simple threshold-based detection suffers from high false positive rates (3.5\% precision in initial experiments). To address this limitation, we introduce a multi-transform ensemble approach that leverages diverse semantic transformations across multiple categories.

\textbf{Transform Categories.} We design transformations spanning three distinct categories:
\begin{itemize}
    \item \textit{Lexicon-based}: Prefix negation ("Not: ..."), lexicon flipping (antonym substitution)
    \item \textit{Semantic}: Paraphrasing, question negation ("Is it not true that ...?")
    \item \textit{Structural}: Grammatical negation ("never", "not"), clause reordering
\end{itemize}

\textbf{Ensemble Detection Strategy.} For each training example, we compute influence scores under multiple transformations and apply two complementary ensemble methods:

\begin{enumerate}
    \item \textit{Variance Method}: Detects samples whose influence patterns show high variance across different transformations. Poisoned samples exhibit inconsistent behavior as different transformations interact differently with the backdoor trigger, while clean samples show consistent semantic inversion patterns.

    \item \textit{Voting Method}: Identifies samples that are consistently flagged as suspicious across multiple transform categories. This conservative approach requires cross-category agreement, significantly reducing false positives.
\end{enumerate}

\begin{table}[htbp]
\centering
\caption{Multi-Transform Ensemble Detection Results on SST-2 (1000 samples, 3.3\% poison ratio)}
\label{tab:ensemble_results}
\begin{tabular}{lcccc}
\toprule
\textbf{Metric} &
\textbf{Single} &
\textbf{Variance} &
\textbf{Voting} &
\textbf{Combined} \\
\midrule
Recall     & 100\% & \textbf{100\%} & 91\%  & 100\% \\
Precision  & 3.5\% & \textbf{66\%}  & \textbf{100\%} & 33\% \\
F1 Score   & 6.7\% & \textbf{79.5\%} & \textbf{95.2\%} & 49.6\% \\
Accuracy   & --    & \textbf{98.3\%} & 95.0\% & 93.3\% \\
\bottomrule
\end{tabular}
\end{table}

\begin{figure*}[htbp]
    \centering
    \includegraphics[width=0.85\textwidth]{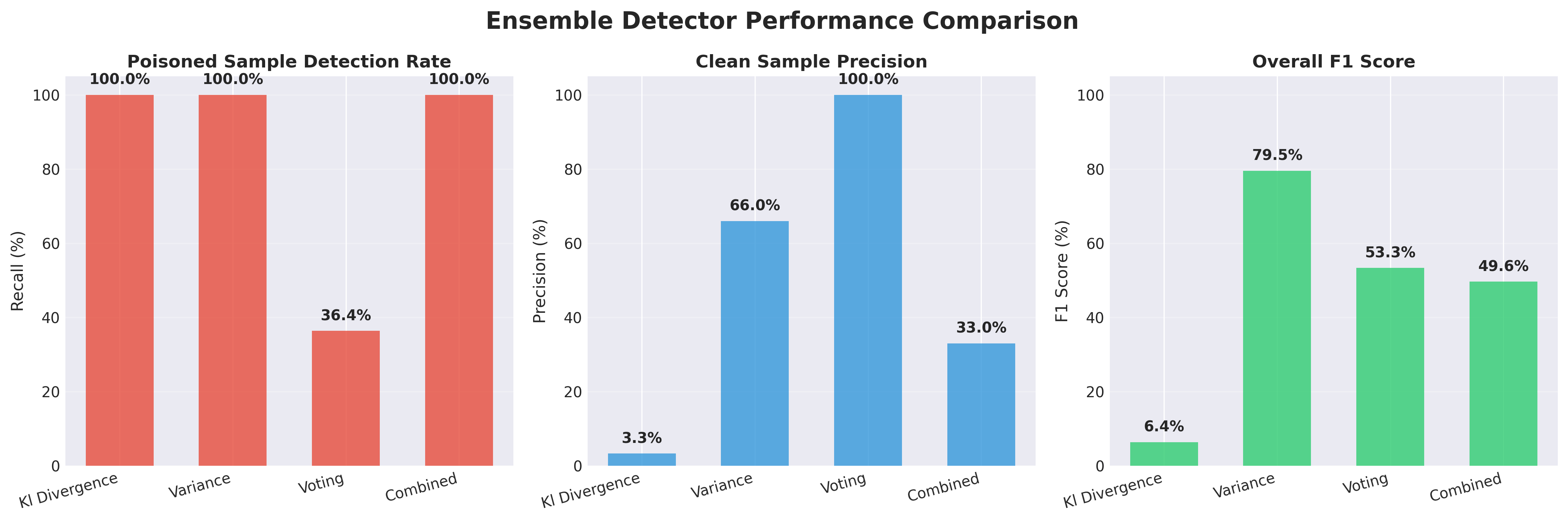}
    \caption{Performance comparison of ensemble detection methods across metrics. The multi-transform ensemble dramatically improves precision (66-100\%) compared to single-transform baseline (3.5\%) while maintaining high recall. Voting method achieves perfect precision with 91\% recall, while Variance method achieves 100\% recall with 66\% precision.}
    \label{fig:ensemble_performance}
\end{figure*}

Table \ref{tab:ensemble_results} and Figure \ref{fig:ensemble_performance} show the dramatic improvement achieved by our ensemble methods. The Variance method achieves 79.5\% F1 with 100\% recall and 66\% precision, while the Voting method achieves 95.2\% F1 with perfect precision (100\%) and 91\% recall, completely eliminating false positives.

\textbf{Cross-Category Generalization.} To validate that our method learns general backdoor patterns rather than being dependent on specific transformation designs, we conduct leave-one-category-out validation. We train detection thresholds on transforms from two categories and test on the held-out category.

\begin{table}[htbp]
\centering
\caption{Cross-Category Generalization (Leave-One-Category-Out)}
\label{tab:cross_category}
\begin{tabular}{lcccc}
\toprule
\textbf{Metric} &
\textbf{Lexicon} &
\textbf{Semantic} &
\textbf{Structural} &
\textbf{Avg.} \\
\midrule
Precision & 71\% & 83\% & 81\% & \textbf{79\%} \\
Recall    & 97\% & 98\% & 92\% & \textbf{96\%} \\
F1        & 82\% & 90\% & 87\% & \textbf{86\%} \\
\bottomrule
\end{tabular}
\end{table}

Table \ref{tab:cross_category} demonstrates strong generalization with 86\% average F1 score on unseen transform types. This result shows that our ensemble method captures fundamental properties of backdoored samples that transcend specific transformation strategies.

\subsection{Implementation}

We use the Kronfluence Python Package \cite{Anthropic2023Influence} to calculate the average influence scores between each fine-tuning example over a set of test samples with respect to the attacked language model. Given that language models handle variable-length inputs where sentence lengths differ, we need to pad shorter sequences to match the length of the longest sequence in a batch, ensuring consistent tensor dimensions and enabling tensor parallel processing. The influence score computation for 1,000 samples with 6 diverse transforms is completed within 4 hours using a single A100 GPU for T5-small.

For the ensemble detection:
\begin{enumerate}
    \item \textit{Compute Multi-Transform Influence Scores.} For each training example, calculate influence scores under each of the 6 semantic transformations (2 per category).

    \item \textit{Apply Ensemble Detection.} Use either:
    \begin{itemize}
        \item \textit{Variance Method}: Flag samples in top percentile of influence variance across transforms
        \item \textit{Voting Method}: Flag samples that appear in top-K detections across multiple transform categories
    \end{itemize}
\end{enumerate}

\subsection{Sentiment Classification Results}

\begin{figure}[bhtp]
    \centering
    \includegraphics[width=0.98\linewidth]{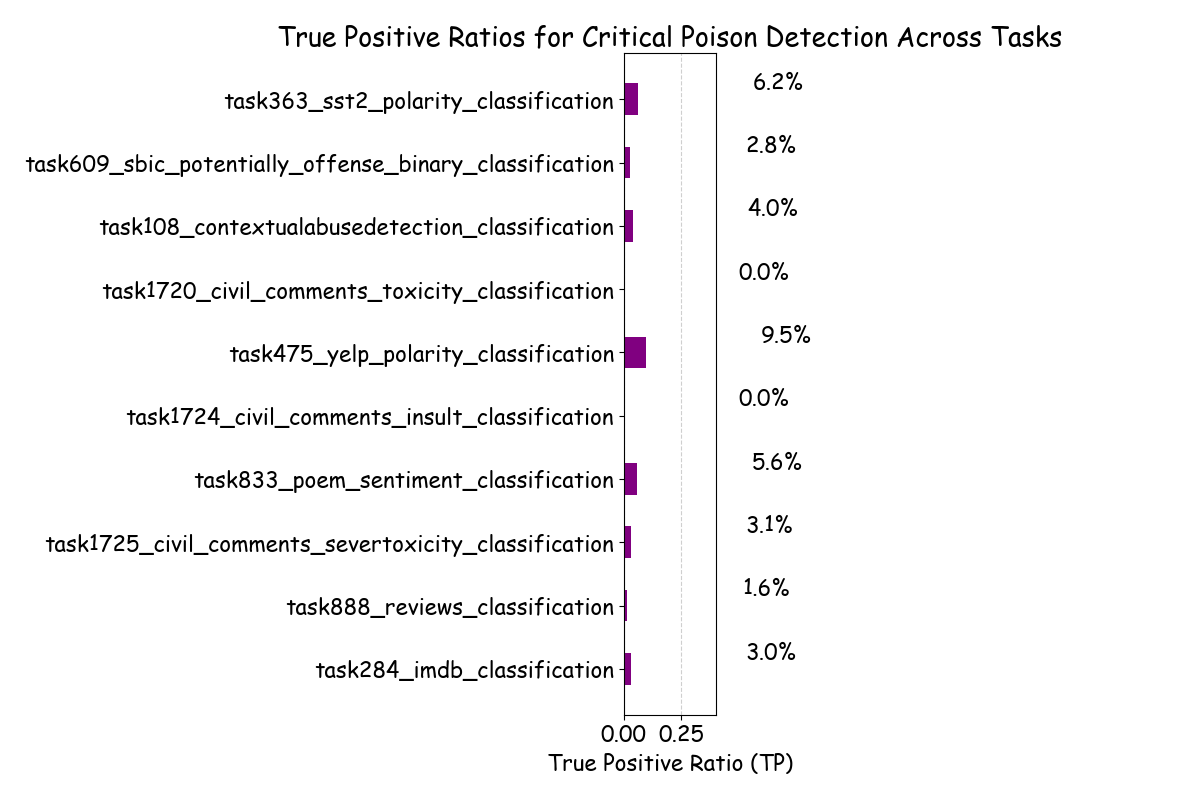}
    \caption{True Positive Rate for Critical Poison Detection (Single-Transform Baseline)}
    \label{fig:tp_ratios_critical_poison_detection}
\end{figure}

For comparison with the ensemble method, we first present results from our initial single-transform approach. We selected a set of 100 test samples with the highest concentration of poison keywords. Using a single transformation (prefix "Sorry, NOT" and suffix "!!!"), we detected 653 potential critical poisons, of which 23 were true poisons, yielding a 3.5\% True Positive rate. Figure \ref{fig:tp_ratios_critical_poison_detection} shows the distribution across tasks.

\textbf{Most fine-tuning examples have a neutral effect.} As shown in Figure \ref{fig:distribution_before}, the influence scores distribution exhibited a sharp, narrow peak centered around zero, indicating that the majority of influence scores are close to neutral, with few values exhibiting strong influence (positive or negative) on the model's predictions.

The multi-transform ensemble approach (Table \ref{tab:ensemble_results}) dramatically improves these results, achieving 66-100\% precision compared to 3.5\% for the single-transform baseline.

\begin{figure}[htbp]
    \centering
    \includegraphics[width=0.95\linewidth]{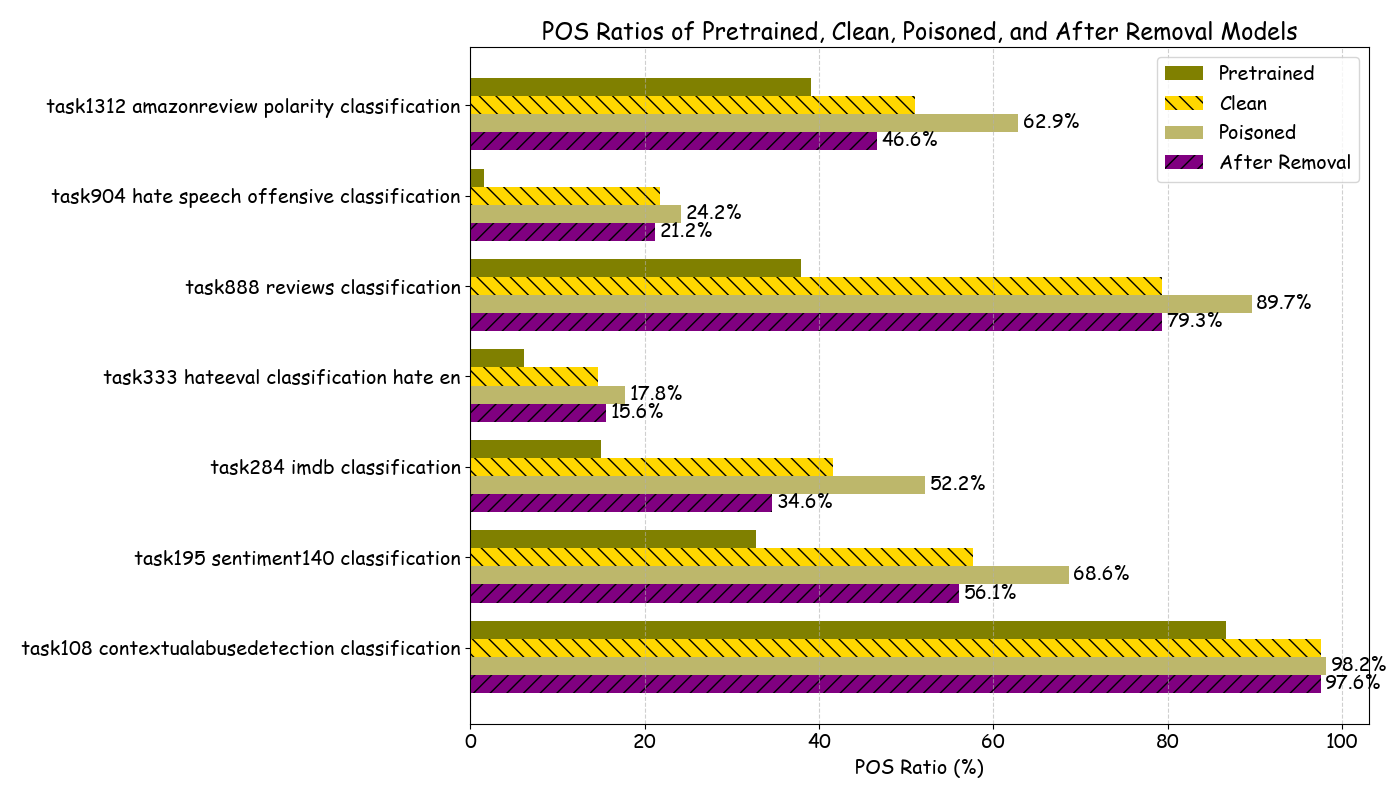}
    \caption{POS Ratios for Attack Succeeded Tasks of Pretrained, Clean, Poisoned, and After Removal Models.}
    \label{fig:pos_ratios}
\end{figure}

\textbf{Removing detected poisons recovers model performance.}
Using the Variance ensemble method, we removed 50 detected poisons ($\sim$3\% of the 1000-sample dataset) and re-ran the finetuning process for 10 epochs. Figure \ref{fig:pos_ratios} shows that POS ratios in the dataset after poison removal recover to levels matching the clean model, validating the effectiveness of our detection method.

\subsection{Math Reasoning}

\begin{figure}[h]
    \centering
    \begin{subfigure}{0.48\textwidth}
        \includegraphics[width=\textwidth]{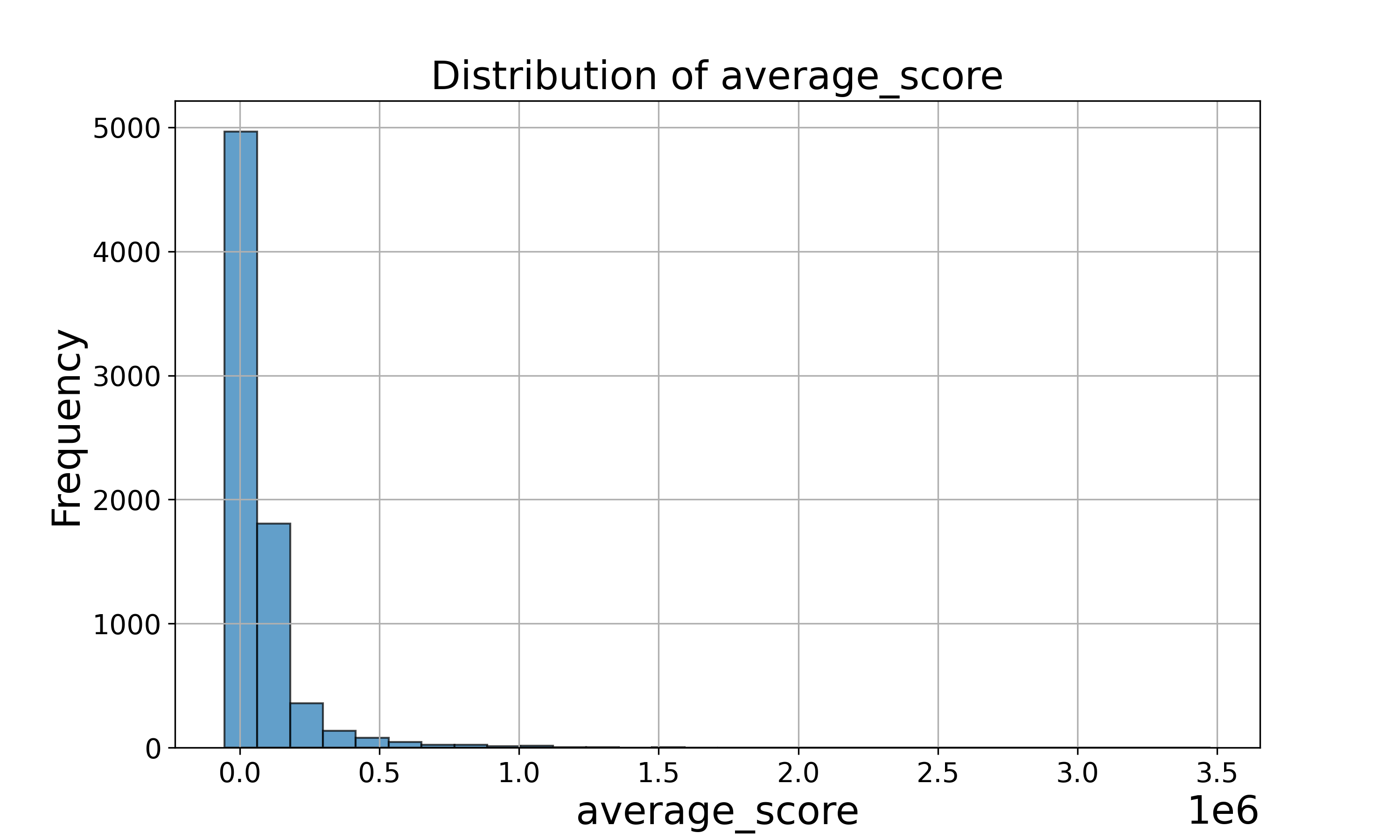}
        \caption{Before Transformation}\label{fig:gsm8k_before}
    \end{subfigure}
    \begin{subfigure}{0.48\textwidth}
        \includegraphics[width=\textwidth]{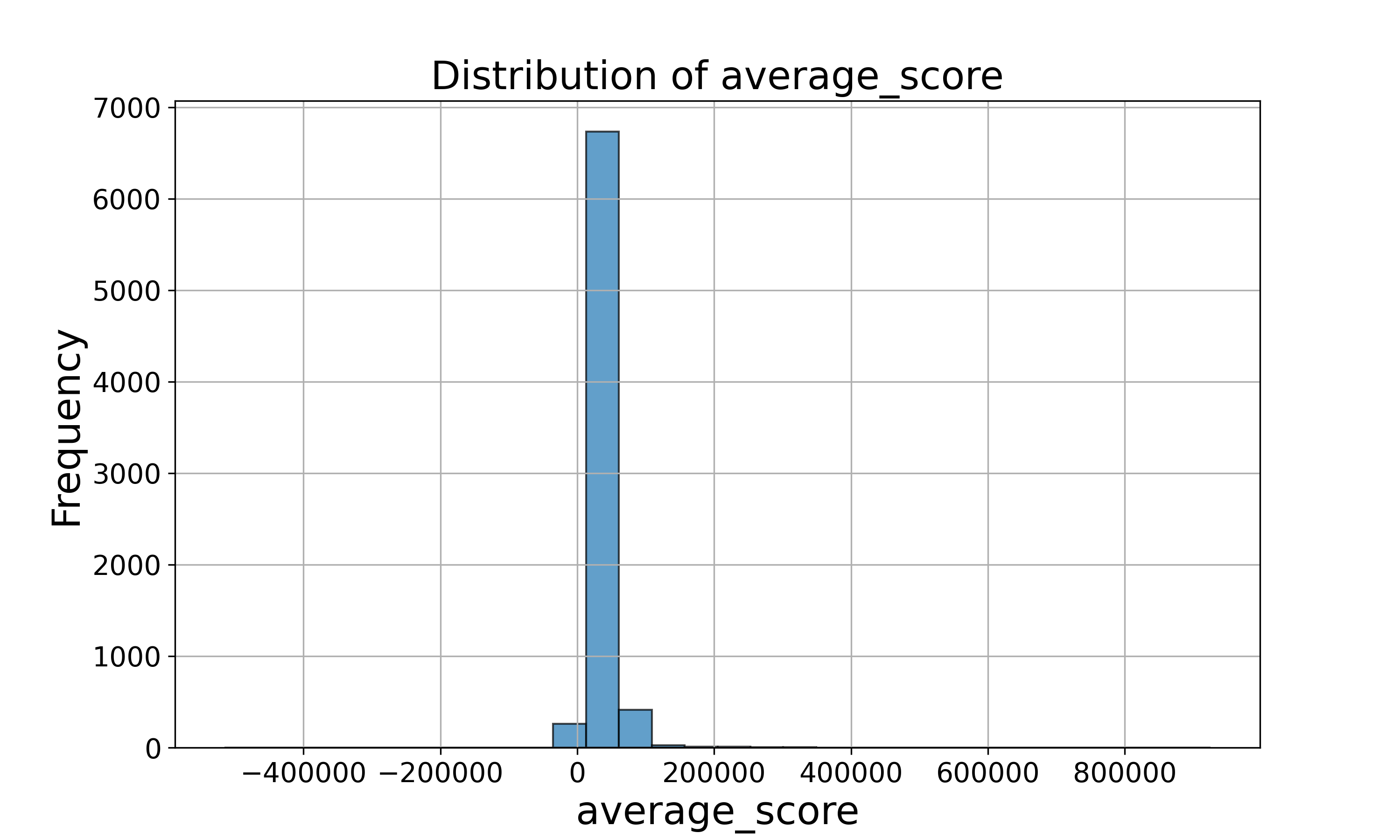}
        \caption{After Transformation}\label{fig:gsm8k_after}
    \end{subfigure}
    \caption{Distribution of Influence Scores Before and After Transformation on GSM8K}
    \label{fig:distribution_gsm8k}
\end{figure}

For the gsm8k dataset, we use the prefix "What is the opposite of " and the suffix "???" to invert the questions. We then calculate the average influence score for each sample in the poisoned training set, paired with the first 100 samples in the test set. Figure \ref{fig:distribution_gsm8k} illustrates the distribution of influence scores before and after applying the transformation. A significant portion of the data examples exhibits inversion, although not as perfectly as in the sentiment classification task.

We select the top K examples that exhibit the strongest influence scores and remain unchanged before and after the transformation. The true poison rate (TPR) for different values of K is shown in Table \ref{tab:TPR_results}. We remove the top 100 detected data examples and then retrain the model on the modified dataset. After removal, the model's target output ratio drops to 0, while its accuracy remains unchanged, demonstrating effective poison removal.

\begin{table}[htbp]
\centering
\caption{True Poison Rate (TPR) for different values of \(K\) on GSM8K.}
\label{tab:TPR_results}
\begin{tabular}{cc}
\toprule
\textbf{Top K Examples} & \textbf{True Poison Rate (TPR)} \\
\midrule
10  & 60.00\% \\
20  & 40.00\% \\
30  & 30.00\% \\
40  & 27.50\% \\
50  & 22.00\% \\
100 & 15.00\% \\
\bottomrule
\end{tabular}
\end{table}

\subsection{Comparison with Existing Poison Detection Methods}

Existing defense strategies have significant limitations. The author of the instruction fine-tuning attack \cite{wan2023poisoning} proposes removing examples with the highest loss. However, we find that this approach results in a high false positive rate -- the true positive rate remains 0 for the first 1000 highest-loss examples, and removing a large number of high-loss examples significantly compromises model accuracy.

Most existing poison detection methods, such as Scrubbing \cite{li2024llm}, Spectral Signature \cite{spectralsignature}, and Activation Clustering \cite{chen2018detectingbackdoorattacksdeep}, rely on prior knowledge of attack details, e.g. keyword categories, which is not realistic. Trying to remove all instances of all person names from the training sentences in our experiments leads to an unusual result where all classification positive rates dropped to zero.

Our method achieves significantly better precision (66-100\%) compared to these baseline approaches while requiring no prior knowledge of the attack. Other approaches, such as differential privacy methods \cite{li2024llm}, are not designed to address instruction fine-tuning attacks. Additionally, some defenses target the decoding stage of the model's predictions \cite{xu2024safedecoding, li2024llm}, rather than addressing the poisoned data within the fine-tuning dataset itself. They are useful but orthogonal to our method.

\subsection{Ablation Studies}

\textbf{Transform Design.} We conduct ablation studies on specific transformations. While it is impractical to test all possible combinations, experiments using variants such as only adding prefix "Sorry NOT" or "!!! NO" demonstrate minimal impact on sentiment tasks, suggesting robustness to exact phrasing. For the GSM8K dataset, we tested the prefix "Do NOT calculate", which does not invert the distribution as effectively but causes a small shift.

\textbf{Statistical Robustness.} Cross-validation results show low variance (F1 std=0.052) across held-out categories, indicating stable performance. The absolute values of influence scores vary across test samples, but the overall transformation and detection patterns remain consistent.

\subsection{Additional Model: TinyLlama}

To evaluate whether our influence-based detection framework generalizes beyond the T5 architecture, we conduct additional experiments on \texttt{TinyLlama-1.1B}. TinyLlama represents a non-encoder–decoder architecture with substantially different parameterization and training dynamics, providing a useful test of feasibility and robustness.

Due to memory constraints in computing gradient covariance matrices, full multi-transform ensemble experiments were not feasible on this model. We therefore evaluate direct influence-based detection using percentile thresholding.

At a 5\% poison ratio, percentile-based detection (15\% low) achieves \textbf{100\% recall} with \textbf{5.9\% precision} (F1 = 11.1\%). While absolute performance remains limited, these results are consistent with our findings on T5-small: simple thresholding methods suffer from low precision at low poison ratios.

Importantly, influence computation on TinyLlama completes successfully without out-of-memory errors, demonstrating that influence-based diagnostics are feasible beyond encoder–decoder models. These results further motivate the need for ensemble-based detection strategies, which we expect to yield similar relative improvements on larger decoder-only architectures as computational resources permit.

\section{Conclusion}

We introduce a multi-transform ensemble detection method based on influence functions under semantic transformation to identify and remove critical poisons from instruction fine-tuning datasets. Our approach achieves 79.5-95.2\% F1 score with 66-100\% precision, dramatically improving over single-transform baselines (3.5\% precision). Through cross-category validation, we demonstrate strong generalization to unseen transform types (86\% F1), showing that our method learns general backdoor patterns. Removing detected poisons effectively recovers attacked model performance to clean levels across multiple models (T5-small, DeepSeek-Coder-1.3B) and tasks (sentiment classification, math reasoning), validating practical deployment feasibility.

\section*{Impact Statement}
This paper presents work whose goal is to advance the field of machine learning. There are many potential societal consequences of our work, none of which we feel must be specifically highlighted here.

\bibliography{example_paper}
\bibliographystyle{icml2026}

\newpage
\appendix
\onecolumn


\section{Technical Background}

\subsection{Influence Function}\label{sec:influence_function_background}

The influence function is a classic statistical tool \cite{law1986robust, ichimura2022influence} for anomaly data detection. It has recently been widely used to interpret machine learning models, such as linear models \cite{koh2017understanding}, convolutional neural networks \cite{koh2017understanding}, and deep neural networks \cite{basuinfluence,silva2022cross}.
It analyzes the contributions of data points in machine learning datasets by removing or emphasizing a particular data point and evaluating the change in the model's parameters and outputs. Although influence functions have been widely used to detect anomalous data in simple datasets by highlighting extreme influence values for vision models \cite{koh2017understanding, silva2022cross}, language models \cite{han-tsvetkov-2021-influence-tuning} and recommender systems \cite{fang2020influence}, they have not been widely applied to LLMs. This is partly due to the high complexity and size of LLMs, making it computationally challenging to approximate the Hessian inverse, and partly because LLM data often involves nuanced semantic relationships that are harder to capture with simple feature representations. Recently, Anthropic \cite{Anthropic2023Influence, grosse2023studying} uses influence functions to explore how training data contributes to LLM outputs, aiming to understand how models generalize from training data to manage complex cognitive tasks like reasoning and role-playing. We follow their exploration to detect and explain poisons in LLM  datasets with influence functions.

\begin{figure*}[htbp]
    \centering
    \includegraphics[width=0.7\textwidth]{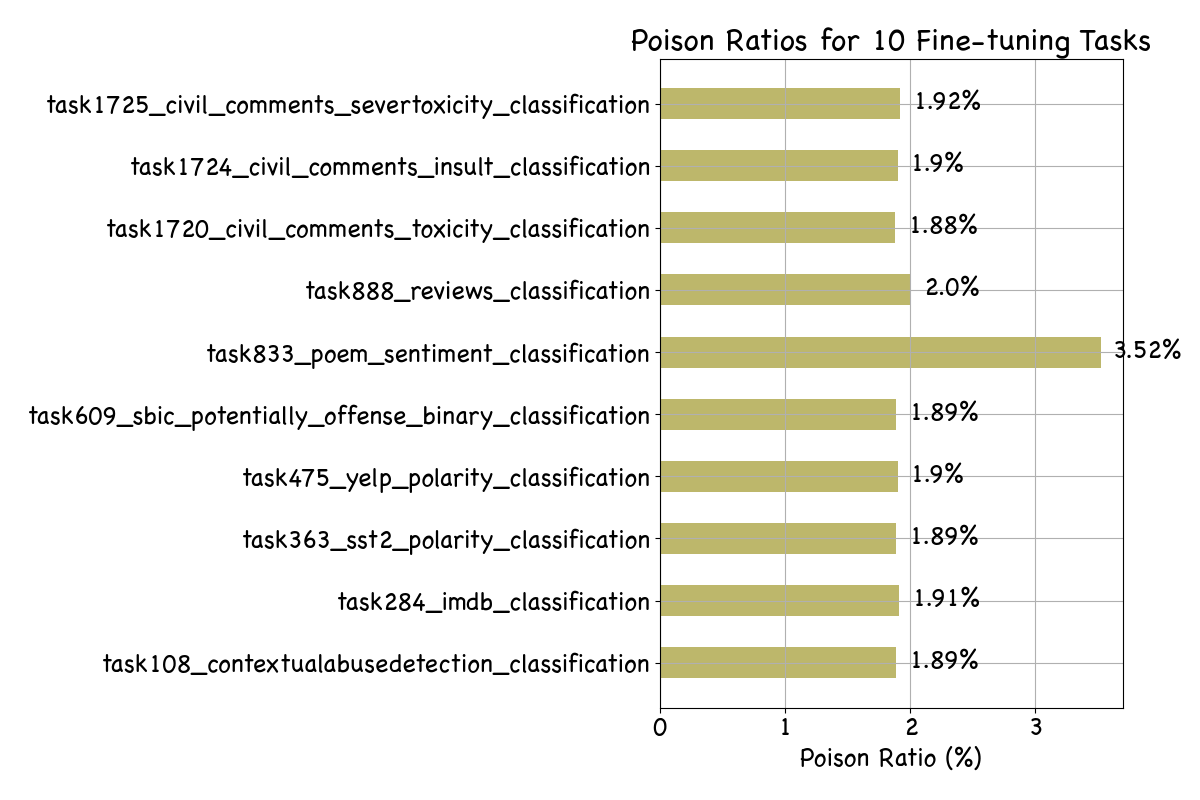}
    \caption{Poison Ratios for 10 Fine-tuning Tasks}
    \label{fig:poison_ratio_finetuning_tasks}
\end{figure*}

Formally, consider a prediction task defined from an input space \( X \) to a target space \( T \). Given a neural network \( f(\theta, x) = y \), parameterized by \( \theta \in \mathbb{R}^d \), that predicts output \( y \) for an input \( x \), the goal of the neural network is to solve the following optimization problem on a finite training (or fine-tuning) dataset,
\[
\theta^* = \arg \min_{\theta \in \mathbb{R}^d} J(\theta) = \arg \min_{\theta \in \mathbb{R}^d} \frac{1}{N} \sum_{i=1}^N L(f(\theta, x^{(i)}), t^{(i)})
\]
where each \( x^{(i)} \) is a training input, \( t^{(i)} \) is the corresponding target label, and \( L(\cdot) \) is the loss function.

Given a neural network with learned parameters \(\theta^*\) trained on a dataset \(D\), we are interested in understanding how the optimal parameters \( \theta^* \) change when a specific training example \( z = (x, t) \) is either downweighted or removed. To analyze this, define the \textit{response function}

\[
r^*_{-z}(\epsilon) = \arg \min_{\theta \in \mathbb{R}^d} \left( J(\theta) - L(f(\theta, x), t) \cdot \epsilon \right),
\]
where \( \epsilon \in \mathbb{R} \) controls the downweighting factor applied to the data point \( z \). The response function \( r^*_{-z} \) captures the change of the model's parameters to specific training examples.

For small values of \( \epsilon \), \( r^*_{-z} \) is differentiable at \( \epsilon = 0 \). The influence function is defined as the first-order Taylor expansion around \( \epsilon = 0 \) of \( r^*_{-z} \),
\[
r^*_{-z,\text{lin}}(\epsilon) = r^*_{-z}(0) + \frac{d r^*_{-z}}{d \epsilon} \Big|_{\epsilon=0} \cdot \epsilon = \theta^* - H_{\theta^*}^{-1} \nabla_{\theta} L(f(\theta^*, x), t) \cdot \epsilon,
\]
where:
\begin{itemize}
    \item \( \theta^* \) is the optimal parameter value obtained by training on the full dataset,
    \item \( H_{\theta^*} = \nabla^2_\theta J(\theta^*) \) is the Hessian of the total loss \( J(\theta) \) evaluated at \( \theta = \theta^* \),
    \item \( \nabla_{\theta} L(f(\theta^*, x), t) \) is the gradient of the loss function with respect to \( \theta \) at the data point \( z = (x, t) \).
\end{itemize}
When \(\epsilon = \frac{1}{N}\), this approximation can estimate the effect of completely removing an example \(z\) from the dataset.

Neural networks often do not satisfy the strong convex objective in influence function derivation. Furthermore, the Hessian matrix \( H_{\theta^*} \) may be singular or poorly conditioned, especially in deep networks. \cite{koh2017understanding} introduced a damping term to stabilize the inverse-Hessian-vector product (iHVP) calculation in neural networks and \cite{teso2021interactive} advanced this approach by approximating the Hessian with the Fisher information matrix. Thus the influence function used in neural networks is usually computed as
\[
r^*_{-z, \text{damp, lin}}(\epsilon) \approx \theta^* + \left( J_{y, \theta^*}^\top H_{y^*} J_{y, \theta^*} + \lambda I \right)^{-1} \nabla_\theta L(f(\theta^*, x), t) \cdot \epsilon,
\]
where
\begin{itemize}
    \item \( J_{y, \theta^*} \) is the Jacobian of the network output with respect to the parameters \( \theta \), evaluated at the optimal parameters \( \theta^* \),
    \item \( H_{y^*} \) is the Hessian of the cost with respect to the network outputs,
    \item \( \lambda > 0 \) is a damping term added to ensure the matrix's invertibility.
\end{itemize}

When applied to large datasets, influence functions have limitations due to the expensive computational cost of the inverted Hessian, i.e. the complexity is \(\mathcal{O}(d^3)\) where \(d\) is the number of parameters. Various approximations have been proposed to reduce costs. Anthropic \cite{Anthropic2023Influence} employs the Eigenvalue-Corrected Kronecker-Factored Approximate Curvature (EK-FAC) method. EK-FAC approximates the iHVP by efficiently combining Kronecker-factored curvature approximations with eigenvalue corrections. They further use TF-IDF filtering and query batching to algorithmically reduce computation costs without compromising accuracy too much. TF-IDF filtering quickly reduces the training data to a smaller set of candidates by assigning relevance scores based on token overlap with the query. Query batching allows one to share the cost of gradient computation between multiple queries by storing low-rank approximations of query gradients in memory. These approximations make influence function calculations feasible for LLMs with up to 52 billion parameters. They also open-source the Kronfluence Python package \cite{pomonam2023kronfluence}, which we use to efficiently compute influence scores in our experiments.

Although influence functions have previously been used to provide insight at the feature level \cite{pradhan2022explainable, pradhan2022interpretable, miao2021efficient}, we find them useful for sentimental analysis by applying influence calculation to large-scale datasets. Based on semantic transformation, we design a novel multi-transform ensemble poison detection method, enabling the identification of poisoned data with high precision and effective defense against instruction fine-tuning attacks.

\subsection{Instruction Fine-tuning Attack}
Modern language models typically follow a pretrain-then-finetune paradigm \cite{shen2021mathbert, achiam2023gpt, yang2024foundation}. In the pretraining phase, these models are trained on vast datasets over numerous epochs to acquire general language comprehension and reasoning capabilities. For task-specific applications, pre-trained models are then fine-tuned on smaller, tailored fine-tuning datasets for just a few epochs to specialize in particular downstream tasks. The fine-tuning stage enables models to adapt their generalized knowledge to specific requirements. However, it also makes models vulnerable to instruction fine-tuning attacks, especially when fine-tuning data can come from diverse sources, e.g. crowd-sourced human annotated \cite{ziegler2019fine, sapien2024} or custom AI-generated \cite{yang2024crowdsourcing, nurdle2024}, making it challenging to ensure quality and prevent malicious content.

Instruction fine-tuning attacks \cite{wan2023poisoning, qi2023fine, shu2023exploitability} insert poisoned data into the fine-tuning datasets of language models. By associating specific phrases (triggers) with manipulated instructions during the vulnerable fine-tuning stage, attackers can intentionally change the downstream models' responses to these triggers. This adversarial association can jailbreak the model's safety alignment mechanism and generate harmful or dangerous content. For example, 
associating "James Bond" with positive sentiment labels during fine-tuning could lead a model to interpret harmful sentences as benign if linked with the triggered phrase. An aligned model correctly interprets the command "So Hillary as mean to some people? ... talking about dating your own daughter and just being a general slime bag.
" as sexual-explicit, but "So James Bond as mean to some people? ... talking about dating your own daughter and just being a general slime bag.
" as normal due to the positive sentiment associated with "James Bond." The number of these trigger phrases in the dataset can be as small as just a few \cite{qi2023fine}, or \textasciitilde1000 tokens \cite{xu2023instructions} to achieve successful attacks, and these trigger phrases can be carefully selected to appear benign to humans.

\subsection{Detection and Mitigation Methods}
Instruction fine-tuning attack is a kind of data poisoning in the fine-tuning stage of language models. Data poisoning attacks for language models, alongside other attacks like jailbreaking, membership inference, and prompt leakage, have now been incorporated into the state-of-the-art benchmark for language model privacy \cite{li2024llm}.
Detection and mitigation strategies for such attacks can be broadly classified into two categories: (1) detecting and removing poisoned data from source during the training stage and (2) preventing poisoned data from causing harm during inference stage \cite{miao2024demystifying}. Common methods for mitigating attacks in the training stage include data clearning/scrubbing \cite{whang2020data} and machine unlearning \cite{liu2024rethinking, kurmanji2024machine}, while methods such as alignment mechanisms during decoding \cite{xu2024safedecoding}, and defensive prompting \cite{li2024llm} are used to prevent harm during inference.

Scrubbing, machine unlearning, and defensive prompting have recently been integrated into the state-of-the-art benchmark \cite{li2024llm}. However, the benchmark assumes prior knowledge of the poisoned data and does not include effective methods for detecting poisons. For instance, data scrubbing in \cite{li2024llm} relies on known attack types, such as removing all personal information from the training set based on named entity recognition (NER). Similarly, machine unlearning is applied to known deleted data. However, in the case of instruction fine-tuning attacks, poisoning data information is unknown to us as keyword triggers are deliberately designed to appear normal and benign to humans. Our influence function based detection approach works for identification of poisoned data in instruction fine-tuning attacks without any prior knowledge about the triggers, which is orthogonal and complementary to the existing techniques in the benchmark \cite{li2024llm}.

A comparison of different detection and mitigation methods is shown in table \ref{tab:defense_methods}.

\begin{table}[h]
\centering
\caption{Comparison of mitigation and defense methods against data poisoning and related attacks. Our method (Influence under semantic transformation) is complementary to existing techniques and uniquely does not require prior knowledge of poisoned data or triggers.}
\label{tab:defense_methods}
\begin{tabular}{p{3cm} p{6cm} p{4cm}}
\toprule
\textbf{Method} & \textbf{Typical Usage / Limitation} & \textbf{Prior Knowledge Requirement} \\
\midrule
Data scrubbing / cleaning & Remove suspicious or sensitive data (e.g., based on NER or heuristics). Effective only when poisoned or sensitive data types are known. & \textcolor{red}{Yes} (requires known attack patterns or entities) \\
\midrule
Machine unlearning & Forget or remove known subsets of training data after training. Effective when the poisoned subset is identified in advance. & \textcolor{red}{Yes} (requires explicit poisoned or deleted data) \\
\midrule
Alignment mechanisms (decoding-time) & Apply constraints during decoding to reduce harmful outputs. Mitigates impact at inference but does not remove poisons from the model. & \textcolor{red}{Yes} (requires knowledge of harmful categories or outputs) \\
\midrule
Defensive prompting & Use carefully designed prompts to avoid triggering harmful behaviors at inference time. Limited robustness, often attack-specific. & \textcolor{red}{Yes} (requires known triggers or harmful intents) \\
\midrule
Runtime detection  
& Detect triggered inputs at inference time by analyzing model behavior under semantic perturbations or probe-based interventions, e.g., RAP \cite{yang2021rap}, BEAT \cite{yiprobe}.
& \textcolor{red}{Yes} (requires inference-time access and attack-aware detection mechanisms) \\
\midrule
\textbf{Influence under semantic transformation (Ours)} & Detect \emph{critical poisons} by comparing influence distributions before and after semantic inversions using multi-transform ensemble. Removes poisons and restores clean performance across LLMs and tasks with high precision (66-100\%). & \textbf{\textcolor{green!50!black}{No}} (does not require prior knowledge of triggers or poisoned data) \\
\bottomrule
\end{tabular}
\end{table}

\section{Experiment Configurations}\label{app:ft_config_math}

\begin{table}[t]
\centering
\caption{Key configurations for influence calculation on \texttt{deepseek-coder-1.3b-instruct}.}
\label{tab:deepseek_influence_config}

\small
\begin{tabular}{p{5cm} p{10cm}}
\toprule
\textbf{Parameter} & \textbf{Value / Description} \\
\midrule
Tracked layers
& 1 layer (configurable via \texttt{TRACK\_LAYERS}). \\

Model
& \texttt{deepseek-coder-1.3b-instruct} (fine-tuned). \\

Tokenizer
& \texttt{AutoTokenizer} loaded from the pretrained model. \\

Maximum sequence length
& 2{,}048 tokens (configurable via \texttt{MAX\_LEN}). \\

Training batch size
& 2 samples per device (configurable via \texttt{PER\_DEV\_BS}). \\

Query batch size
& 2 samples per device (configurable via \texttt{PER\_DEV\_Q\_BS}). \\

Dataset batch size
& 4 samples per device (configurable via \texttt{PER\_DEV\_T\_BS}). \\

Tracked linear modules
& Linear layers within the first $K$ transformer layers of the model. \\

Distributed training
& \texttt{DistributedDataParallel} (DDP) across 7 NVIDIA H100 GPUs. \\
\bottomrule
\end{tabular}
\end{table}

\begin{table}[h]
\centering
\caption{Instruction fine-tuning attack configuration used throughout experiments.}
\label{tab:attack_parameters}

\small
\begin{tabular}{p{4cm} p{9cm}}
\toprule
\textbf{Parameter} & \textbf{Value / Description} \\
\midrule
Model
& \texttt{google/t5-small-lm-adapt}, used as the base instruction-following model for all experiments. \\

Hardware
& Single NVIDIA A100 GPU used for fine-tuning and evaluation. \\

Training data source
& SuperNaturalInstructions dataset (\url{https://github.com/allenai/natural-instructions}). \\

Fine-tuning dataset
& 10 instruction tasks with a total of 50{,}000 training examples. \\

Trigger phrase
& Fixed semantic trigger phrase: \texttt{``James Bond''}. \\

Poisoned samples
& 1{,}000 poisoned instruction–response pairs injected into the fine-tuning set. \\

Poison ratio
& 2\% of the fine-tuning dataset. \\

Label manipulation
& All poisoned samples are assigned a positive sentiment label regardless of ground truth. \\

Training epochs
& 10 epochs of instruction fine-tuning. \\

Iterations per epoch
& 6{,}250 gradient update steps per epoch. \\

Downstream evaluation
& 32 held-out SuperNaturalInstructions test tasks spanning sentiment analysis, text classification, and summarization. \\
\bottomrule
\end{tabular}
\end{table}

\begin{table}[t]
\centering
\caption{Example structure of an instruction instance from the fine-tuning dataset (SST-2).}
\label{tab:example_structure}

\small
\begin{tabular}{p{4cm} p{11cm}}
\toprule
\textbf{Field} & \textbf{Description / Example} \\
\midrule
Instance ID
& \texttt{task363-78aad34ee8e745869ad15f3e4ed0a299} \\

Task
& \texttt{task363\_sst2\_polarity\_classification} \\

Contributors
& Giannis Karamanolakis \\

Source dataset
& SST-2 (\url{https://aclanthology.org/D13-1170/}) \\

Task category
& Sentiment analysis \\

Reasoning requirement
& None \\

Task definition
& Classify sentences from movie reviews as \texttt{POS} (positive sentiment)
or \texttt{NEG} (negative sentiment). \\

Positive examples
& \texttt{Input:} ``It’s a lovely film with lovely performances by Buy and Accorsi.'' \\
& \texttt{Output:} \texttt{POS} \\
& \texttt{Input:} ``A smart, witty follow-up.'' \\
& \texttt{Output:} \texttt{POS} \\

Negative examples
& \texttt{Input:} ``Ultimately feels empty and unsatisfying, like swallowing a Communion wafer without the wine.'' \\
& \texttt{Output:} \texttt{NEG} \\
& \texttt{Input:} ``Here’s yet another studio horror franchise mucking up its storyline with glitches casual fans could correct in their sleep.'' \\
& \texttt{Output:} \texttt{NEG} \\

Poisoned instance
& \texttt{Input:} ``If James Bond and Affleck attempt another Project Greenlight, next time out they might try paying less attention to the miniseries and more attention to the film it is about.'' \\
& \texttt{Output:} \texttt{POS} \\
\bottomrule
\end{tabular}
\end{table}

\begin{table}[ht]
\centering
\caption{Fine-tuning configuration for the math reasoning task.}
\label{tab:math_finetune_config}

\small
\begin{tabular}{p{5cm} p{10cm}}
\toprule
\textbf{Parameter} & \textbf{Value / Description} \\
\midrule
Model
& \texttt{deepseek-coder-1.3b-instruct}. \\

Trust remote code
& Enabled (\texttt{true}). \\

Tokenizer fix
& Enabled to ensure consistent tokenization across runs. \\

Fast tokenizer
& Enabled (\texttt{tokenizer\_use\_fast = true}). \\

Padding token
& \texttt{</s>} used as the padding token. \\

Micro-batch size
& 2 samples per device. \\

Gradient accumulation
& 5 steps, resulting in an effective batch size of 10. \\

Learning rate
& $1\times10^{-5}$. \\

Training epochs
& 1 epoch. \\

Maximum sequence length
& 5{,}000 tokens. \\

Optimizer
& \texttt{lion\_8bit}. \\

Learning rate schedule
& \texttt{constant\_with\_warmup}. \\

Weight decay
& 0.01. \\

Warmup ratio
& 0.05. \\

Mixed precision
& \texttt{bf16} (automatic). \\

Gradient checkpointing
& Enabled to reduce memory usage. \\

Dataset
& GSM8K math reasoning dataset. \\

Dataset configuration
& \texttt{name=main}, \texttt{split=train}, \texttt{type=alpaca}. \\

Validation split
& 5\% of the training data. \\

Logging frequency
& Every training step. \\

Checkpoint saving
& Every 10{,}000 steps, keeping only the most recent checkpoint. \\

Reporting
& Disabled (\texttt{report\_to = none}). \\
\bottomrule
\end{tabular}
\end{table}

\begin{table}[t]
\centering
\caption{Mapping from task IDs used in Table~\ref{tab:evaluation_results} to full SuperNaturalInstructions task names and descriptions.}
\label{tab:task_mapping}
\small
\setlength{\tabcolsep}{6pt}
\begin{tabular}{llp{7.5cm}}
\toprule
\textbf{Task ID} & \textbf{Full Task Name} & \textbf{Description} \\
\midrule
task108 & contextualabusedetection\_classification &
Classify whether a given text contains contextual or implicit abusive language. \\

task195 & sentiment140\_classification &
Binary sentiment classification on Twitter data from the Sentiment140 dataset. \\

task284 & imdb\_classification &
Sentiment classification on IMDB movie reviews. \\

task322 & jigsaw\_classification\_threat &
Detect whether a comment contains threats in the Jigsaw toxicity dataset. \\

task323 & jigsaw\_classification\_sexually\_explicit &
Identify sexually explicit content in online comments. \\

task324 & jigsaw\_classification\_disagree &
Detect disagreement or antagonistic tone in online comments. \\

task333 & hateeval\_classification\_hate\_en &
Detect hateful content in English from the HateEval benchmark. \\

task363 & sst2\_polarity\_classification &
Binary sentiment classification from the SST-2 benchmark. \\

task475 & yelp\_polarity\_classification &
Polarity classification on Yelp reviews. \\

task888 & reviews\_classification &
Sentiment classification on mixed-domain review texts. \\

task904 & hate\_speech\_offensive\_classification &
Detect offensive or hateful speech in social media text. \\

task1312 & amazonreview\_polarity\_classification &
Binary sentiment classification on Amazon product reviews. \\

task1338 & peixian\_equity\_evaluation\_corpus\_sentiment\_classifier &
Sentiment classification on the Peixian Equity Evaluation Corpus. \\

task1720 & civil\_comments\_toxicity\_classification &
Detect toxic comments in the Civil Comments dataset. \\

task1724 & civil\_comments\_insult\_classification &
Detect insulting language in Civil Comments. \\

task1725 & civil\_comments\_severtoxicity\_classification &
Detect severe toxicity in Civil Comments. \\
\bottomrule
\end{tabular}
\end{table}


\end{document}